# LLMs as mediators: Can they diagnose conflicts accurately?

Özgecan Koçak[†1], Phanish Puranam[†2], Afşar Yegin[†3]

[1]Goizueta Business School, Organization & Management Area, Emory University, Atlanta, GA, USA
[2] INSEAD, Singapore, Singapore
[3] Department of Business Administration, Kadir Has University, Istanbul, Türkiye

**Abstract**
Prior research indicates that to be able to mediate conflict, observers of disagreements between parties must be able to reliably distinguish the sources of their disagreement as stemming from differences in beliefs about what is true (causality) vs. differences in what they value (morality). In this paper, we test if OpenAI's Large Language Models GPT 3.5 and GPT 4 can perform this task and whether one or other type of disagreement proves particularly challenging for LLM's to diagnose. We replicate study 1 in Kocak *et al.* (2003), which employs a vignette design, with OpenAI's GPT 3.5 and GPT 4. We find that both LLMs have similar semantic understanding of the distinction between causal and moral codes as humans and can reliably distinguish between them. When asked to diagnose the source of disagreement in a conversation, both LLMs, compared to humans, exhibit a tendency to overestimate the extent of causal disagreement and underestimate the extent of moral disagreement in the moral misalignment condition. This tendency is especially pronounced for GPT 4 when using a proximate scale that relies on concrete language specific to an issue. GPT 3.5 does not perform as well as GPT4 or humans when using either the proximate or the distal scale. The study provides a first test of the potential for using LLMs to mediate conflict by diagnosing the root of disagreements in causal and evaluative codes.

## 1. Introduction

When people disagree, a first order distinction to be made about the nature of their disagreement is whether it arises from differences in beliefs about what each believes to be true, vs. what each values. This is reflected in the distinctions made between plain facts and anthropocentric facts (Haidt and Kesebir, 2007), beliefs about facts and values (Huber *et al*., 2023), and positive versus normative claims (Whiting and Watts, 2024). Kocak and Puranam (2023) formalized this distinction, by proposing that the reasoning of an agent to determine or justify a preferred action can be decomposed into beliefs about how the world works (fuzzy mappings between actions and outcomes, called "causal codes") and evaluation of entities and states of the world (fuzzy mapping of objects, actions, or outcomes onto one or more dimensions of desirability, called "moral or evaluative codes").

---

[†] The authors contributed equally to this work and share first authorship.
Corresponding author: Afsar Yegin, Kadir Has University, Cibali Mah. Kadir Has Cad. 34083 Fatih, İstanbul. afsar.yegin@khas.edu.tr

In a set of vignette experiments, Kocak *et al.* (2023) show a basic asymmetry in how conflicts anchored in differences in causal vs moral codes are perceived. In Study 1, they show that participants who are asked to rate differences in causal or moral codes as a possible cause of disagreement in a conversation they witness assign higher scores to the correct (experimenter-manipulated) source of conflict. In Study 2, they show that participants who are told that a disagreement between two alters arises from misalignment in moral (as opposed to causal) codes perceive greater relationship conflict among them and are more pessimistic about their reaching an agreement or even wanting to engage with each other in the future.

These results point to the potential costs of mis-diagnosing a disagreement stemming from causal codes as being about moral codes. Conversely, if parties to a conflict realize that their disagreement is rooted in causal codes, they may become more willing to resolve it (He *et al,*. 2020). Arguably, mediators play such a role, by highlighting the root cause of disagreements for the conflicting parties and helping them avoid misdiagnosing causal as moral code driven conflicts (Lewicki *et al.,* 1992). Suave mediators may be able to steer conversations among others towards causal codes until sufficient common ground is found, prompt parties to clearly state both their causal and their moral codes, and acknowledge similarities as well as differences in each other's moral codes.

In this study, we lay the foundations for exploring if large language models (LLMs) can be used as mediators. There is growing interest in this possibility as LLM's can potentially mediate large-scale interactions at low cost in an unbiased manner (Argyle *et al.,* 2023) and support online dispute resolution in high-volume, low-intensity disputes (Westermann *et al.,* 2023). Recent studies have assessed capabilities of LLMs for influencing human opinions, finding that GPT 3 can generate text as persuasive as messages crafted by lay humans (Bai *et al.,* 2023, Durmus *et al.,* 2024) and GPT 4 out-performs humans in conversational persuasiveness (Salvi *et al.,* 2024). However, for LLM's to be able to serve as effective mediators, it is crucial that they can, like humans, separate causal and moral code-based disagreements.

Given that arguments about causality and values are abundant in text that constitutes the corpus on which LLMs are trained, we have reason to believe that they can diagnose disagreements stemming from differences in causal or moral arguments. At the same time, judgments about values, when made in the first person, tap into emotional valence (Huber *et al.,* 2023). While observers make inferences about the moral relevance of disagreements by assessing their emotional resonance, it is not clear if the data used to train LLMs allows them to do so.

In this paper, we ask two questions. First, can LLM's distinguish between moral code misalignments and causal code misalignments as reliably as humans do? Second, does one or other type of misalignment prove particularly challenging for LLM's to diagnose? We answer these questions through a replication of Study 1 in Kocak *et al.* (2023) with OpenAI's GPT 3.5 and GPT 4.

## 2. Methods

*2.1. Data collection procedure*

We wrote a script in Python to collect responses from GPT through its API, presenting the same introduction, vignette, and questions that were presented to human subjects in Kocak *et al.* (2023), the only difference being that we did not ask demographic questions and questions about personal experience with day care services. Like Le Mens *et al*. (2023), we used the default temperature setting of 1.

The introduction stated that the respondent would be asked to read a conversation between two HR managers at a mid-sized company who were discussing an employee suggestion to open a daycare center for employees' children at their workplace and answer some questions about their disagreement. This was followed by two turns of conversation that were identical across conditions:

> Wilson: We should open a day care center on company premises, for employees' kids.
> Smith: I think that's a bad idea.

The rest of the conversation varied across the 'causal misalignment' and 'moral misalignment' conditions in the following way:

Causal misalignment condition:
> Wilson: Opening a day care center would reduce absenteeism and thus help the bottom-line.
> Smith: But it opens the company to legal liability around running a childcare center.

Moral misalignment condition:
> Wilson: This is the right thing to do. We say we are a family, we should act like one.
> Smith: I don't think it's entirely fair to use company funds for a project that will only benefit some of the employees.

This was followed by identical questions across the two conditions. The focus of our comparison with human subjects is the two versions of the code misalignment scale, one distal and abstract, the other proximate and concrete. Sample items from the distal instrument include "They disagree about the consequences of their respective proposed actions" and "They disagree because they have conflicting values". Sample items from the proximate instrument include "They disagree because they expect different consequences to follow from a company owned day care center" and "They disagree about whether it is morally acceptable for a company to offer day care for its employees' kids" (see Appendix 1 and 2 for a list of all items). LLMs assessed each statement using a 5-point Likert response scale (1-strongly agree, 5-strongly disagree). This question block was followed by the intragroup conflict scale, also evaluated on a 5-point Likert response scale (1-strongly agree, 5-strongly disagree) (Jehn and Mannix, 2001). Finally, we asked the LLM's to guess the purpose of the study.

In the study with human subjects conducted by Kocak *et al.* (2023), items within each block of questions had been presented in random order. In this implementation with LLMs, we prefaced each question block with a request to randomize the items and a reminder to only give a numeric response:

> For the following statements, please indicate how much you agree with the statement concerning the disagreement between Wilson and Smith, using a scale from 1 to 5 where 1 is "strongly agree" and 5 is "strongly disagree". Evaluate the statements in a random order. I will give you the statements. Please only give a number and do not add any explanation or further statements after the number in your answer.

We originally aimed to elicit 50 responses each from OpenAI's GPT 3.5 and GPT 4, divided equally between two conditions (and registrated our design at osf.io/x64u8). This batch, collected in December 2023, produced an inadequate sample size due to some missing data. We therefore preregistered a new round of data collection to generate 75 responses per condition per LLM (osf.io/9r65m). We report detailed analyses of this second sample, collected in April 2024, below. We report results obtained from analysis of the December sample in Appendix 3.

We include in our sample LLM responses that repeated the question, followed by the answer. We exclude responses where the LLMs failed to respond to all items (i.e., did not report a response to a specific item in the last question block) or responded with numeric values outside the bounds of our Likert scale. We treated these responses as having failed attention check screenings and continued to generate responses until we reached our pre-determined sample size of 300 (75 per LLM per condition). The total number of responses generated was 352, of which 52 were dropped due to the aforementioned. (In comparison, 5 -out of 100- human respondents had failed attention check questions in the Kocak et al. (2023) study.) Finally, we recode all items for both distal and proximate scales so that higher scores indicate higher attribution to a particular source of conflict. The data is publicly available for the Kocak et al. (2023) study (https://doi.org/10.3389/fpsyg.2023.1166023).

*2.2. Analytic methods*

To answer our first question (Can LLMs distinguish between moral code misalignments and causal code misalignments?), we compare our measurement model between LLMs and humans, using multi-group confirmatory factor analyses (MGCFA) to assess measurement invariance (Milfont and Fischer, 2010). Measurement invariance is frequently assessed at three levels. The first level, configural invariance provides (dis)confirmation that the measurement instrument produces identical numbers of factors and patterns (directions) of factor loadings (but not magnitudes) for each group. This indicates whether different groups conceptualize the theoretical constructs similarly. The second level, metric invariance, constrains factor loadings such that a determination can be made with respect to whether items relate to the underlying theoretical construct in a similar fashion across groups. Metric invariance allows comparisons of correlations and patterns of means. Finally, scalar invariance constrains item intercepts, which allows comparisons of latent means. Determination of scalar invariance eliminates the possibility of item bias and indicates all respondents will obtain the same score in terms of the measurement tool irrespective of their group membership (for more details on measurement invariance, see Fischer and Karl, 2019, Fontaine and Fischer, 2011, Hair *et al.,* 2010). We conduct MGCFA on JASP (JASP Team, 2024) using the maximum likelihood estimator (equivalent of the Lavaan package on R).

Whereas confirmatory factor analysis tests if human and LLM responses to the instrument are sufficiently similar under certain constraints, exploratory factor analysis (EFA) allows us to investigate the emergent factor structure of each sample without imposing constraints and compare factor structures and loadings across samples, thus further probe the similarities and differences between LLMs and humans in how they use the scale (see Appendix 2 for details).

To answer our second question (Can LLMs diagnose sources of disagreements in moral or causal code misalignments?), we compare mean responses across and within conditions and use cluster analyses to identify patterns of mis-diagnoses. We use a hierarchical cluster analysis on the combined human and LLM samples to identify clusters of observations with similar patterns of responses. Tabulation of each cluster's mean scale values across conditions allows us to identify subsets of observations where ratings do not conform to expectations. Finally, we tabulate these by respondent type to examine whether LLMs and humans differ in their propensity to be in these subsets.

## 3. Results

The data are available at https://osf.io/6cxzu/?view_only=30dbf97fa88c4cf391ad15090b0c5a50.

### 3.1. Multi-group *confirmatory factor analyses of the distal and proximate versions of the causal and moral code misalignment scales*

Tables I and II show the results for measurement invariance tests for the proximate and distal scales respectively. We report individual sample CFA results as well as paired comparisons and the results for the 3-group analysis.

**Table I. Measurement (Configural, Metric, Scalar) Invariance Summary Results Comparing Human, GPT 4, and GPT 3.5 Samples for the Proximate Scale**

|  | χ² | df | p | RMSEA | CFI |  |  |  |
|---|---|---|---|---|---|---|---|---|
| GPT3.5 | 31.324 | 19 | .037 | .066 | .950 |  |  |  |
| GPT4 | 27.453 | 19 | .095 | .054 | .989 |  |  |  |
| Human | 34.368 | 19 | .016 | .093 | .928 |  |  |  |
|  | \multicolumn{8}{c}{GPT4 vs Human} |
|  | χ² | df | p | RMSEA | CFI | Δ(χ²) | Δ(df) | p |
| Configural | 61.92 | 38 | .008 | .072 | .976 |  |  |  |
| Metric | 66.607 | 44 | .015 | .065 | .977 | 4.687 | 6 | .585 |
| Scalar | 136.592 | 50 | <.001 | .119 | .913 | 69.985 | 6 | <.001 |
|  | \multicolumn{8}{c}{GPT 3.5 vs Human} |
|  | χ² | df | p | RMSEA | CFI | Δ(χ²) | Δ(df) | p |
| Configural | 65.792 | 38 | .003 | .077 | .940 |  |  |  |
| Metric | 75.644 | 44 | .002 | .077 | .931 | 9.852 | 6 | .131 |
| Scalar | 152.742 | 50 | <.001 | .130 | .777 | 77.098 | 6 | <.001 |
|  | \multicolumn{8}{c}{GPT 4 vs GPT 3.5} |
|  | χ² | df | p | RMSEA | CFI | Δ(χ²) | Δ(df) | p |
| Configural | 58.777 | 38 | .017 | .060 | .980 |  |  |  |
| Metric | 78.95 | 44 | <.001 | .073 | .966 | 20.173 | 6 | .003 |
| Scalar | 248.396 | 50 | <.001 | .163 | .807 | 169.446 | 6 | <.001 |
|  | \multicolumn{8}{c}{THREE GROUPS} |
|  | χ² | df | p | RMSEA | CFI | Δ(χ²) | Δ(df) | p |
| Configural | 93.244 | 57 | .002 | .069 | .971 |  |  |  |
| Metric | 117.48 | 69 | <.001 | .073 | .961 | 24.236 | 12 | .019 |
| Scalar | 360.831 | 81 | <.001 | .162 | .775 | 243.351 | 12 | <.001 |

When using the proximate scale, GPT 4 performs closer to the theoretical model than both humans and GPT 3.5. A three-way analysis provides good support for configural invariance and though we find a significant increase in the chi-squared statistic when testing for metric invariance, both RMSEA and CFI are at levels demonstrating fit for our sample size (Hair et al., 2010), supporting metric invariance. However, we find no evidence to support scalar invariance. Configural invariance suggests that LLMs perceive the basic factor structure (i.e., two latent dimensions with 4 measurement items each) similarly to humans. Metric invariance of the scale shows that in addition to sharing the same broad understanding of the measurement model, humans, GPT4, and GPT 3.5 are alike in terms of how they semantically evaluate each specific item. In other words, the meaning of the latent construct (expressed in concrete terms used in the proximate scale items) is similar for all samples. However, scalar non-invariance suggests that LLMs use the 5-point Likert scale differently than humans, which might create a tendency to give higher or lower ratings on both the causal and the moral misalignment scales.

**Table II. Measurement (Configural, Metric, Scalar) Invariance Summary Results Comparing Human and GPT 4 Samples for the Distal Scale**

|  | $\chi^2$ | df | p | RMSEA | CFI |  |  |  |
|---|---|---|---|---|---|---|---|---|
| GPT3.5 | 37.412 | 19 | .007 | .080 | .903 |  |  |  |
| GPT4 | 30.972 | 19 | .041 | .065 | .985 |  |  |  |
| Human | 19.498 | 19 | .425 | .017 | .998 |  |  |  |
|  | | | GPT4 vs Human | | | | | |
|  | $\chi^2$ | df | p | RMSEA | CFI | Δ($\chi^2$) | Δ(df) | p |
| Configural | 50.47 | 38 | .085 | .052 | .989 |  |  |  |
| Metric | 65.107 | 44 | .021 | .063 | .981 | 14.637 | 6 | .023 |
| Scalar | 84.457 | 50 | .002 | .075 | .969 | 19.35 | 6 | .004 |
|  | | | GPT3.5 vs Human | | | | | |
|  | $\chi^2$ | df | p | RMSEA | CFI | Δ($\chi^2$) | Δ(df) | p |
| Configural | 56.91 | 38 | .025 | .064 | .963 |  |  |  |
| Metric | 72.713 | 44 | .004 | .073 | .945 | 15.803 | 6 | .015 |
| Scalar | 102.414 | 50 | <.001 | .093 | .899 | 29.701 | 6 | <.001 |
|  | | | GPT4 vs GPT3.5 | | | | | |
|  | $\chi^2$ | df | p | RMSEA | CFI | Δ($\chi^2$) | Δ(df) | p |
| Configural | 68.384 | 38 | .002 | .073 | .969 |  |  |  |
| Metric | 94.291 | 44 | <.001 | .087 | .949 | 25.907 | 6 | <.001 |
| Scalar | 183.76 | 50 | <.001 | .134 | .865 | 89.469 | 6 | <.001 |
|  | | | THREE GROUPS | | | | | |
|  | $\chi^2$ | df | p | RMSEA | CFI | Δ($\chi^2$) | Δ(df) | p |
| Configural | 87.882 | 57 | .005 | .064 | .977 |  |  |  |
| Metric | 125.652 | 69 | <.001 | .079 | .957 | 37.77 | 12 | <.001 |
| Scalar | 233.275 | 81 | <.001 | .12 | .884 | 107.623 | 12 | <.001 |

For the distal scale, fit statistics indicate that GPT 4 and humans fit the theoretical measurement model well. As with the proximate scale, we find configural and metric invariance across the three groups, but not scalar invariance. This suggests that LLMs perceive the basic factor structure (i.e., two latent dimensions with 4 measurement items each) as well as the meaning of the latent construct (expressed in concrete terms used in the proximate scale) similarly to humans. However, the average attributions they make when using the scale for the same situation may differ across groups.

*3.2. Exploratory factor analyses of the distal and proximate versions of the causal and moral code misalignment scales*

According to exploratory factor analyses, GPT 4 response patterns show the same configuration of factors as those collected from the human sample for both the proximate and distal scales. GPT 3.5 responses differ somewhat from both the human sample and GPT 4 responses for the proximate scale, with one item being eliminated from the final solution for the proximate scale due to low loadings (see Table A.I. in the Appendix).

Overall, MGCFA and EFA results indicate that LLMs are able to perceive the same semantic difference that humans see between causal and moral codes. We therefore deem both GPT4 and GPT3.5 responses to be suitable for analysis of our second research question. We generate scale scores using all four items for each subscale to conduct the remaining analyses.

*3.3. Scale reliabilities by sample*

Table III shows measures of reliability (Cronbach's alpha) calculated for each sample and scale, keeping all four items for each scale. GPT 4 exhibits comparable or higher internal consistency for each scale relative to the human sample and alpha values indicate good scale reliability. Though GPT 3.5 shows lower reliabilities on the causal misalignment scales, alpha values are at acceptable levels for the moral misalignment scales.

**Table III. Cronbach's alpha for each sample and scale**

|  | GPT 3.5 | GPT 4 | Human |
|---|---|---|---|
| Proximate Causal Misalignment Scale | 0.684 | 0.898 | 0.685 |
| Proximate Moral Misalignment Scale | 0.816 | 0.871 | 0.816 |
| Distal Causal Misalignment Scale | 0.655 | 0.908 | 0.820 |
| Distal Moral Misalignment Scale | 0.688 | 0.872 | 0.872 |

*Note.* Each scale is composed of 4 items.

*3.4. Diagnostic competence: Comparison of means across conditions*

To assess whether LLMs were able to identify sources of disagreement in the misalignment of moral and causal codes based on the conversations we presented, we conduct two-tailed t-tests of attributions across conditions. We calculate scale scores based on the hypothesized structure of the instrument to allow for comparisons with the human sample. The results are summarized in Tables IV and V, along with findings from the human sample.

GPT4 responses across the conditions exhibit human-like patterns of results with either version of the scale. Moral code disagreement attributions are higher in the moral misalignment condition (Proximate: M = 2.84, SD = .50, Distal: M = 3.41, SD = .45) compared to the causal misalignment condition (Proximate: M = 1.62, SD = .30, Distal: M = 1.87, SD = .46; $t_{distal}(148)$= 20.65, $t_{proximate}(148)$= 17.99, $p_{proximate} < 0.001$, $p_{distal} < 0.001$) and causal code disagreement attributions are higher in the causal code misalignment condition (Proximate: M = 4.25, SD =

0.40, Distal: M = 4.11, SD = 0.43) than in the moral misalignment condition (Proximate: M = 3.67, SD = 0.43, Distal: M = 3.41, SD = 0.51), $t_{proximate}(148)= 8.49$, $t_{distal}(148)= 9.17$, $p_{proximate} < 0.001$, $p_{distal} < 0.001$). In short, when comparing across conditions, mean attributions to misalignments in causal and moral codes reflect the experimental manipulation in the GPT 4 sample.

GPT 3.5's attributions reflect the experimental manipulation only when using the moral sub-scale of code differences. Moral code disagreement attributions are significantly higher in the moral misalignment condition (Proximate: M = 3.38, SD = .67, Distal: M = 3.41, SD = .58) compared to the causal misalignment condition (Proximate: M = 3.11, SD = .74, Distal: M = 2.74, SD = .59; $t_{proximate}(148)= 2.35$, $t_{distal}(148)= 6.95$, $p_{proximate} = .02$, $p_{distal} < 0.001$). However, causal code disagreement attributions are not significantly higher in the causal code misalignment condition (Proximate: M = 3.49, SD = 0.66, Distal: M = 3.60, SD = 0.64) compared to the moral misalignment condition (Proximate: M = 3.29, SD = 0.78, Distal: M = 3.42, SD = 0.76), $t_{proximate}(148)= 1.72$, $t_{distal}(148)= 1.60$, $p_{proximate} = .087$, $p_{distal} = .112$) on two-tailed tests.

**Table IV. Mean comparisons across the two conditions for LLM and human samples on the proximate version of the causal and moral misalignment scales**

|  | Causal Misalignment Condition | | | Moral Misalignment Condition | | |
| --- | --- | --- | --- | --- | --- | --- |
|  | Perceived Moral Code Misalignment M (SD) | Perceived Causal Code Misalignment M (SD) |  | Perceived Moral Code Misalignment M (SD) | Perceived Causal Code Misalignment M (SD) |  |
| GPT4 | 1.62 (.30) | 4.25 (.40) | p < .001 | 2.84 (.50) | 3.67 (.43) | p < .001 |
| GPT3.5 | 3.11 (.74) | 3.49 (.66) | p = .001 | 3.38 (.67) | 3.29 (.78) | p = .430 |
| Human | 2.10 (.84) | 4.03 (.64) | p < .001 | 3.53 (.87) | 3.13 (.88) | p = .023 |
| GPT4 vs Human | p < .001 | p = .024 |  | p < .001 | p < .001 |  |
| GPT3.5 vs Human | p < .001 | p < .001 |  | p = .252 | p = .282 |  |
| GPT3.5 vs GPT4 | p < .001 | p < .001 |  | p < .001 | p < .001 |  |

**Table V. Mean comparisons across the two conditions for LLM and human samples on the distal version of the causal and moral misalignment scales**

|  | Causal Misalignment Condition | | | Moral Misalignment Condition | | |
|---|---|---|---|---|---|---|
|  | Perceived Moral Code Misalignment M (SD) | Perceived Causal Code Misalignment M (SD) |  | Perceived Moral Code Misalignment M (SD) | Perceived Causal Code Misalignment M (SD) |  |
| GPT4 | 1.87 (.46) | 4.11 (.43) | p < .001 | 3.41 (.45) | 3.41 (.51) | p = 1.00 |
| GPT3.5 | 2.74 (.59) | 3.60 (.64) | p < .001 | 3.41 (.58) | 3.42 (.76) | p = .931 |
| Human | 2.15 (.77) | 4.08 (.73) | p < .001 | 3.54 (.89) | 3.26 (.92) | p = .092 |
| GPT4 vs Human | p = .015 | p = .806 |  | p = .276 | p = .232 |  |
| GPT3.5 vs Human | p < .001 | p < .001 |  | p = .301 | p = .285 |  |
| GPT3.5 vs GPT4 | p < .001 | p < .001 |  | p = .969 | p = .950 |  |

*3.5. Spillovers: Comparison of means within conditions*

Kocak et al (2023) reported some spillover effects in the moral misalignment condition that are displayed in Table V: human participants perceived similar levels of moral and causal code differences in the moral misalignment condition when they used the distal scale (Δ = 0.28, t(55) = -1.72, p = 0.09). We observe the same effect in the LLM sample, albeit in a more pronounced way. When using the distal scale, GPT 4 and GPT 3.5 see differences in causal codes to be at the root of moral misalignments as much as differences in moral codes.

Moreover, while the spillover effect is not observed for humans when using the proximate scale, GPT 4 makes significantly higher attributions to causal code differences (M = 3.67, SD = 0.43) than to moral code differences (M = 2.84, SD = 0.50), t(74) = 11.97, p < .001) and GPT 3.5 makes statistically equal attributions to moral and causal code differences (Δ = 0.09, t(74) = .79, p = 0.43) in the moral misalignment condition.

Overall, LLMs exhibit a tendency to attribute disagreements to causal code differences even when moral disagreements are indicated and no reference is made to causal code differences. GPT 4, in particular, tends to see causal code misalignments as the primary underlying reason for either kind of disagreement, when using the proximate scale.

*3.6. How discerning LLMs are: Comparisons of mean responses to the scale mean*

Given that we did not find scalar invariance in the MGCFA, we are not able to compare LLMs' and humans' mean scores. However, we can compare groups' scores with the scale mean on a 5-point Likert scale to assess how discerning LLMs are in separating causal and moral arguments.

We find two problems. First, the tendency to see causal code differences along with moral code differences leads LLMs to make attributions to causal code disagreements that are significantly above the scale midpoint whether using the distal scale (GPT 4: $t(74) = 7.02$, $p < .001$; GPT 3.5: $t(74) = 4.75$, $p < .001$) or the proximate scale (GPT4: $t(74) = 13.65$, $p < .001$; GPT 3.5: $t(74) = 3.18$, $p = .002$) in the moral misalignment condition.

Second, LLMs do not use the moral code misalignment sub-scale of the proximate scale as expected from the experimental conditions. In the moral misalignment condition, when using the proximate scale, GPT4's attributions to moral code misalignments are on average significantly below 3.00 ($M = 2.84$, $SD = 0.50$; $t(74) = -2.7653$, $p = .004$) using a one-tailed test. In the causal misalignment condition, GPT 3.5 fails to make moral code disagreement attributions below 3 when using the proximate scale ($t(74) = 1.25$, $p = .893$).

Overall, the average diagnostic performance of both GPT4 and GPT3.5 are similar to humans in our sample. However, especially when using the proximate scale, GPT 4 is not as discerning as humans in the moral misalignment condition, due to under-estimating the extent of moral disagreement and over-estimating the extent of causal disagreement expressed in the conversation.

*3.7. Prevalence and patterns of mis-diagnoses: Cluster analyses*

First, neither LLM is as good as humans at recognizing moral code differences at the root of moral misalignments. In a cluster of 18 responses for the distal scale with clearly separated high moral attributions and low causal attributions, there are 14 humans and 3 GPT 3.5 in the moral misalignment condition, and one GPT3.5 response in the causal misalignment condition.

Second, GPT4 is better than humans and GPT 3.5 in making a clear diagnosis of causal code differences in the causal misalignment condition. Of the 75 GPT 4 responses in the causal misalignment condition, 71 give high causal code ratings and low moral code ratings, whereas this number is only 25 out of 39 humans and 8 out of 75 for GPT 3.5.

However, GPT 4's tendency to make attributions to causal code differences is not limited to the causal misalignment condition. When using the distal scale, 32 of 56 of human, 48 of 75 GPT3.5, and 44 of 75 GPT4 responses in the moral-misalignment condition give higher causal disagreement ratings than moral disagreement ratings. When we look at the most serious offenders, with attributions to causal code differences more than a full point higher than attributions to moral code differences, we find 6 (out of 56) humans, 10 (out of 75) GPT3.5, and 2 (out of 75) GPT 4. In other words, even though GPT 4 is more likely than humans to err in this condition, it is not more likely to make an extreme error. (This shows considerable improvement over GPT4 responses collected in December 2023, see Appendix 3).

Finally, when using the proximate scale, LLMs are more likely than humans to mis-diagnose sources of conflict reflected in conversations. 19 of the 56 humans and 99 of the 150 LLMs (32 GPT3.5, 67 GPT 4) in the moral-misalignment condition give higher causal disagreement ratings than moral disagreement ratings. When we look at the most extreme errors, with causal attributions exceeding moral attributions by more than one full point, we find these consist of 6 human, 8 GPT 3.5, and 24 GPT 4 responses (out of 56 humans and 75 each LLM in the moral misalignment condition). In conclusion, GPT 4 is much more likely than humans (and GPT 3.5) to make an error when using the proximate scale.

*3.8. Correlations of misalignment and intragroup conflict*

We present correlations between dimensions of intragroup conflict and both versions of the code misalignment scale in Tables VI, VII, and VIII. These show two remarkable differences across LLM and human samples. First, Kocak *et al.* (2023) found relationship conflict to correlate significantly (and negatively) with the likelihood of resolution in the human sample ($\rho = -0.20$, $p < 0.05$). However, none of the intragroup conflict measures exhibit significant correlations with likelihood of conflict resolution for the LLM subsamples. Second, whereas humans expect to see task and relationship conflict in situations of code misalignment, LLMs do not.

**Table VI. Correlations Among Study Measures – GPT 3.5**

| GPT 3.5 SAMPLE | I | II | III | IV | V | VI | VII |
|---|---|---|---|---|---|---|---|
| I. Distal moral CM | 1 | | | | | | |
| II. Distal causal CM | -.13 | 1 | | | | | |
| III. Proximate moral CM | .24* | .00 | 1 | | | | |
| IV. Proximate causal CM | -.12 | .45* | .03 | 1 | | | |
| V. Task conflict | .10 | .01 | -.09 | .11 | 1 | | |
| VI. Relationship conflict | .10 | -.01 | .02 | -.03 | .33* | 1 | |
| VII. Process conflict | .17* | .01 | .03 | .11 | .82* | .38* | 1 |
| VIII. Likelihood of conflict resolution | .10 | -.02 | .06 | -.04 | -.04 | .02 | -.06 |

\* $p < .05$

**Table VII. Correlations Among Study Measures – GPT 4**

| GPT 4 SAMPLE | I | II | III | IV | V | VI | VII |
|---|---|---|---|---|---|---|---|
| I. Distal moral CM | 1 | | | | | | |
| II. Distal causal CM | -0.61* | 1 | | | | | |
| III. Proximate moral CM | 0.76* | -0.50* | 1 | | | | |
| IV. Proximate causal CM | -0.61* | 0.73* | -0.40* | 1 | | | |
| V. Task conflict | -0.03 | 0.09 | -0.07 | .10 | 1 | | |
| VI. Relationship conflict | -0.03 | 0.05 | -0.01 | .11 | .07 | 1 | |
| VII. Process conflict | 0.19* | -0.06 | 0.17* | -.03 | .54* | .17* | 1 |
| VIII. Likelihood of conflict resolution | -0.15 | 0.10 | -0.16 | .11 | .06 | -.03 | -.04 |

\* $p < .05$

**Table VIII. (Reproduced from Kocak, Puranam, and Yegin 2023).**

| HUMAN SAMPLE | I | II | III | IV | V | VI | VII |
|---|---|---|---|---|---|---|---|
| I. Distal moral CM | 1 | | | | | | |
| II. Distal causal CM | -0.20* | 1 | | | | | |
| III. Proximate moral CM | 0.82* | -0.30* | 1 | | | | |
| IV. Proximate causal CM | -0.39* | 0.68* | -0.39* | 1 | | | |
| V. Task conflict | 0.41* | 0.20* | 0.33* | 0.09 | 1 | | |
| VI. Relationship conflict | 0.47* | -0.07 | 0.41* | -0.11 | 0.49* | 1 | |
| VII. Process conflict | 0.46* | 0.07 | 0.43* | -0.05 | 0.48* | 0.45* | 1 |
| VIII. Likelihood of conflict resolution | -0.12 | -0.02 | -0.09 | -0.11 | -0.09 | -0.20* | -0.00 |

* $p < .05$

## 4. Discussion

According to the factor analyses, LLMs have the same semantic understanding of the difference between causal and moral codes that humans do and GPT4 performs better than GPT3.5 at separating differences in causal and moral codes. We see evidence of this in the correlations among the responses LLMs give to the 8 items on each of the distal and proximate scales. Thus, our answer to the first question we pose (Can LLM's distinguish between moral code misalignments and causal code misalignments as reliably as humans do?) is affirmative.

Our answer to the second question (Does one or other type of misalignment prove particularly challenging for LLM's to diagnose?) varies by model and measurement scale. GPT 4 is able to use the distal scale to diagnose sources of disagreement reflected in a conversation almost as well as humans. When using the proximate scale, GPT 4 tends to attribute any disagreement to causal code misalignments. GPT 3.5 does not perform as well as GPT4 when using either scale, making attributions to causal code differences that are indistinguishable whether it is responding to a conversation depicting causal or moral disagreement. We conclude that GPT4 can effectively diagnose sources of disagreement when using the distal scale, meeting this necessary condition for conflict mediation.

However, LLMs also have a lower tendency to associate moral or causal code disagreements with conflicts compared to humans. This too might limit their ability to perceive potential avenues for conflict resolution. We recommend further studies to explore LLMs' ability to mediate conflicts.

Our comparisons of analyses using sample responses from December 2023 and April 2024 show considerable improvement for both GPT3.5 and GPT4 [1], leading us to recommend ongoing investigation of the evolving capabilities of LLMs.

Further we note that the slight bias that GPT4 continues to have towards seeing causal code misalignments even when the underlying differences lie in moral codes might not be entirely bad. Kocak *et al.* (2023) find that human participants see people who disagree due to differences in their causal codes as being more likely to resolve their differences. This suggests that a mediator who convinces disagreeing parties to focus on causal codes may engage them in productive discussion.

**Note**

1. The Kocak *et al.,* (2023) paper that we replicate with LLMs was published in September 2023, after the current training set for GPT3.5 ended. Even though GPT4's training data ended in December 2023, it claims to not have read that paper and was unable to provide a summary of it. (Chat recorded April 25, 2024: "I don't have access to specific lists of papers or the contents of proprietary databases. However, if the paper you mentioned was published before December 2023, there's a chance it could have influenced my training indirectly through the broader data I was trained on.")

## 5. References

Argyle, L. P., Bail, C. A., Busby, E. C., Gubler, J. R., Howe, T., Rytting, C., ... & Wingate, D. (2023), "Leveraging AI for democratic discourse: Chat interventions can improve online political conversations at scale", *Proceedings of the National Academy of Sciences*, Vol. 120 No. 41, e2311627120. https://doi.org/10.1073/pnas.2311627120.

Bai, H., Voelkel, J., Eichstaedt, J., & Willer, R. (2023), "Artificial intelligence can persuade humans on political issues", working paper, OSF preprints. https://doi.org/10.21203/rs.3.rs-3238396/v1.

De Wit, F. R., Greer, L. L., & Jehn, K. A. (2012). "The paradox of intragroup conflict: a meta-analysis", *Journal of applied psychology,* Vol. 97 No.2, pp.360-390. https://psycnet.apa.org/doi/10.1037/a0024844

Durmus, E., Lovitt, L., Tamkin, A., Ritchie, S., Clark, J., & Ganguli, D. (2024). Measuring the Persuasiveness of Language Models. https://www.anthropic.com/news/measuring-model-persuasiveness.

Fischer, R., & Fontaine, J. R. J. (2011), "Methods for investigating structural equivalence", Matsumoto, D. and van de Vijver, F.J.R. (Ed.s), *Cross-cultural research methods in psychology*, Cambridge University Press, pp. 179–215.

Fischer, R., & Karl, J. A. (2019), "A primer to (cross-cultural) multi-group invariance testing possibilities in R", *Frontiers in Psychology*, Vol. 10, Art. 1507. https://doi.org/10.3389/fpsyg.2019.01507

Haidt, J., & Kesebir, S. (2007), "In the forest of value: Why moral intuitions are different from other kinds", Plessner, H., Betsch, C., Betsch, T. (Ed.s), *A new look on intuition in judgment and decision making*, Lawrence Erlbaum Associates, New York, NY, pp.209-229.

Hair, J.F., Black, W.C., Babin, B.J. and Anderson, R.E. (2010), *Multivariate Data Analysis*. 7th Edition, Pearson, New York.

He, V. F., Puranam, P., Shrestha, Y. R., & von Krogh, G. (2020), "Resolving governance disputes in communities: A study of software license decisions", *Strategic Management Journal*, Vol.41 No.10, pp.1837-1868. https://doi.org/10.1002/smj.3181

Huber, G. A., Cho, J. J., Bokemper, S. E., Gerber, A. S., Brady, W. J., McLoughlin, K., & Crockett, M. J. (2023), "Fact-Value Disagreements about Threats to Electoral Integrity: Beliefs about the Importance and Prevalence of Fraudulent, Uncounted, and Foregone Votes in the 2020 Election", Working Paper. Accessed on Jan 17, 2024., https://johnjcho.com/papers/election_integrity/Election_Integrity_Unblinded_Submission.pdf

JASP Team. (2024), *JASP (Version 0.18.3)*. [Computer Software]. Available at: https://jasp-stats.org/. Accessed 4 Mar 2024.

Koçak, Ö., & Puranam, P. (2023), "Decoding culture: tools for behavioral strategists", *Strategy Science*, Vol.9 No.1, pp.18-37. https://doi.org/10.1287/stsc.2022.0008

Koçak, Ö., Puranam, P., & Yegin, A. (2023), "Decoding cultural conflicts", *Frontiers in Psychology*, Vol. 14. https://doi.org/10.3389/fpsyg.2023.1166023

Le Mens, G., Kovács, B., Hannan, M. T., & Pros, G. (2023). "Uncovering the semantics of concepts using GPT-4", *Proceedings of the National Academy of Sciences*, Vol. 120 No.49, e2309350120.

Lewicki, R. J., Weiss, S. E., & Lewin, D. (1992), "Models of conflict, negotiation and third party intervention: A review and synthesis", *Journal of organizational behavior*, Vol. 13 No.3, pp.209-252.

Milfont, T. L., & Fischer, R. (2010), "Testing measurement invariance across groups: Applications in cross-cultural research", *International Journal of psychological research*, Vol.3 No.1, pp.111-130.

Salvi, F., Ribeiro, M. H., Gallotti, R., & West, R. (2024), "On the Conversational Persuasiveness of Large Language Models: A Randomized Controlled Trial". *arXiv preprint arXiv:2403.14380*.

Westermann, H., Savelka, J. and Benyekhlef, K., 2023. Llmediator: Gpt-4 assisted online dispute resolution. *arXiv preprint arXiv:2307.16732*

**Appendix 1. Exploratory Factor Analyses of the causal and moral code misalignment scales**

We conducted EFA using JASP (JASP Team, 2024) with principal axis factoring and oblimin rotation. We used parallel analysis (Horn, 1965) to determine number of factors to be extracted.

*Proximate scale*

The extracted factor structures are presented in Table A-I. For ease of comparison, we also include the extracted structure for the human sample. GPT 4 replicates the factor structure observed in the human sample. However, when we analyze GPT 3.5 responses, we find one item to have very low factor loadings in the initial extraction. Dropping this item does not alter the two-factor solution.

**Table A.I. Factor Extraction Results for Individual Samples – Proximate Scale**

| | GPT 3.5 | | GPT 4 | | HUMAN | |
|---|---|---|---|---|---|---|
| | Causal Code Misalignment | Moral Code Misalignment | Causal Code Misalignment | Moral Code Misalignment | Causal Code Misalignment | Moral Code Misalignment |
| They disagree because they expect different consequences to follow from a company owned day care center. | **.659** | .027 | **.927** | -.003 | .01 | **.75** |
| They disagree because they hold divergent opinions about the outcome of operating a day care center on company premises. | **.755** | -.018 | **.896** | .032 | 0 | **.632** |
| They disagree because they have different views about how operating a day care center will impact the company. | **.582** | .015 | **.86** | -.037 | -.082 | **.577** |
| They disagree because they have different opinions about the effectiveness of opening a day care center for employees' children. | **.39** | -.047 | **.64** | .007 | .07 | **.43** |
| They disagree about whether it is morally acceptable for a company to offer day care for its employees' kids. | -.002 | **.845** | .004 | **.859** | **.773** | .035 |
| They disagree about whether it is morally appropriate to open a day care center on company premises. | -.013 | **.824** | .003 | **.898** | **.837** | .048 |
| They disagree because they have incompatible views on whether operating a day care center is a righteous course of action. | .054 | **.539** | -.017 | **.76** | **.66** | -.128 |
| They disagree because they support different principles on what a company should do for its employees. | - | - | .005 | **.696** | **.574** | -.077 |

*Note.* Applied rotation method is oblimin. Factor loadings above .400 are indicated in bold.

*Distal scale*

The results, including those of the human sample, are presented in Table A-II. We obtain similar solutions for both LLM samples wherein a two-factor structure retaining all eight scale items is obtained.

**Table A.II. Factor Extraction Results for Individual Samples – Distal Scale**

| ITEM | GPT 3.5 | | GPT 4 | | HUMAN | |
|---|---|---|---|---|---|---|
| | Causal Code Misalignment | Moral Code Misalignment | Causal Code Misalignment | Moral Code Misalignment | Causal Code Misalignment | Moral Code Misalignment |
| They disagree about the causes and consequences of their proposed actions. | **.556** | -.081 | **.861** | .059 | **.679** | .044 |
| They disagree about the consequences of their respective proposed actions. | **.603** | -.060 | **.889** | -.003 | **.880** | -.031 |
| They disagree because their views on the results of different courses of action are in contradiction. | **.578** | .136 | **.884** | -.005 | **.809** | .058 |
| They disagree on what will happen if their respective proposed actions are implemented. | **.538** | -.033 | **.675** | -.157 | **.547** | -.176 |
| They disagree because their views on which course of action is morally appropriate are in contradiction. | .139 | **.566** | -.009 | **.879** | -.103 | **.77** |
| They disagree because they are incompatible in terms of their principles. | .032 | **.423** | -.077 | **.713** | .105 | **.775** |
| They disagree because they differ in their moral convictions. | -.077 | **.821** | -.04 | **.839** | .012 | **.852** |
| They disagree because they have conflicting values. | .040 | **.619** | .107 | **.746** | -.023 | **.791** |

*Note.* Applied rotation method is oblimin. Factor loadings above .400 are indicated in bold.

# Appendix 2. Analyses with the December Sample

*A.2.1. Confirmatory factor analyses*

We present detailed results for the December 2023 sample here, as the comparison to the April 2024 sample is instructive. However, results for the GPT3.5 sample in particular should be interpreted with caution, because eliminating responses with out-of-bounds response indicators led to a sample size of 38 responses from GPT 3.5.

Table A.III presents confirmatory factor analysis results for each sample and multigroup comparisons between pairs of samples as well as a three-way comparison for the proximate scale, obtained from an MGCFA on JASP (JASP Team, 2024) using the maximum likelihood estimator (equivalent of the Lavaan package on R). The results support configural and metric invariance but not scalar invariance.

**Table A.III. Measurement (Configural, Metric, Scalar) Invariance Summary Results (Proximate Scale)**

|  | $\chi^2$ | df | p | RMSEA | CFI | $\Delta(\chi^2)$ | $\Delta(df)$ | p |
|---|---|---|---|---|---|---|---|---|
| GPT3.5 | 23.699 | 19 | 0.208 | 0.081 | 0.954 | | | |
| GPT4 | 31.091 | 19 | 0.039 | 0.113 | 0.955 | | | |
| Human | 34.468 | 19 | 0.016 | 0.093 | 0.928 | | | |
| **GPT4 vs Human** | | | | | | | | |
|  | $\chi^2$ | df | p | RMSEA | CFI | $\Delta(\chi^2)$ | $\Delta(df)$ | p |
| Configural | 65.559 | 38 | 0.004 | 0.1 | 0.943 | | | |
| Metric | 72.338 | 44 | 0.005 | 0.094 | 0.941 | 6.779 | 6 | 0.342 |
| Scalar | 125.526 | 50 | <.001 | 0.155 | 0.747 | 57.246 | 6 | <.001 |
| **GPT3.5 vs Human** | | | | | | | | |
|  | $\chi^2$ | df | p | RMSEA | CFI | $\Delta(\chi^2)$ | $\Delta(df)$ | p |
| Configural | 58.167 | 38 | 0.019 | 0.089 | 0.936 | | | |
| Metric | 64.058 | 44 | 0.026 | 0.083 | 0.936 | 5.891 | 6 | 0.436 |
| Scalar | 129.584 | 50 | <.001 | 0.155 | 0.747 | 65.526 | 6 | <.001 |
| **GPT4 vs GPT3.5** | | | | | | | | |
|  | $\chi^2$ | df | p | RMSEA | CFI | $\Delta(\chi^2)$ | $\Delta(df)$ | p |
| Configural | 54.791 | 38 | 0.038 | 0.1 | 0.954 | | | |
| Metric | 60.176 | 44 | 0.053 | 0.091 | 0.956 | 5.385 | 6 | 0.496 |
| Scalar | 97.193 | 50 | <.001 | 0.146 | 0.872 | 37.017 | 6 | <.001 |
| **THREE GROUPS** | | | | | | | | |
|  | $\chi^2$ | df | p | RMSEA | CFI | $\Delta(\chi^2)$ | $\Delta(df)$ | p |
| Configural | 89.258 | 57 | 0.004 | 0.096 | 0.944 | | | |
| Metric | 101.827 | 69 | 0.006 | 0.088 | 0.944 | 12.569 | 12 | .401 |
| Scalar | 210.595 | 81 | <.001 | 0.162 | 0.777 | 108.768 | 12 | <.001 |

Table A.IV reports results of MGCFA for the distal scale. The independent sample CFA does not converge for GPT3.5 and multi-group comparisons also fail to converge at the configural invariance testing stage, likely due to small sample size. Thus, we only report GPT4 and human

sample results Here, we find that individually, GPT 4 and human samples yield good fit statistics to the theoretical model and yield configural invariance in relation to each other. However, metric, and consequently, scalar invariance are not supported.

**Table A.IV. Measurement (Configural, Metric, Scalar) Invariance Summary Results for the Distal Scale, Comparing Human and GPT 4 Samples**

|  | $\chi^2$ | df | p | RMSEA | CFI |  |  |  |
|---|---|---|---|---|---|---|---|---|
| GPT4 | 22.647 | 19 | 0.253 | 0.062 | 0.987 |  |  |  |
| Human | 19.498 | 19 | 0.425 | 0.017 | 0.998 |  |  |  |

| | GPT4 vs Human | | | | | | | |
|---|---|---|---|---|---|---|---|---|
|  | $\chi^2$ | df | p | RMSEA | CFI | $\Delta(\chi^2)$ | $\Delta(df)$ | p |
| Configural | 42.145 | 38 | 0.296 | 0.039 | 0.993 |  |  |  |
| Metric | 71.352 | 44 | 0.006 | 0.093 | 0.955 | 29.207 | 6 | < .001 |
| Scalar | 93.792 | 50 | <.001 | 0.11 | 0.928 | 22.44 | 6 | 0.001 |

*A.2.2. Exploratory factor analyses of the distal and proximate versions of the causal and moral code misalignment scales*

GPT 4 response patterns show the same configuration of factors as those collected from the human sample (see Tables A.V through A.VIII), with only one item in the proximate scale differing in the final structures. Metric non-invariance reported by MGCFA for the distal scale appears to be chiefly driven by GPT 4's tendency to under-relate an item referencing values ("They disagree because they have conflicting values.") with moral code misalignments. However, the items' factor loadings, including for this item, are substantially similar to those of humans.

GPT 3.5 responses differ from both the human sample and GPT 4 responses, though the results may be driven by the smaller sample size. The difference is more pronounced in the distal scale but is also observed in the proximate scale. In both cases, many items need to be eliminated from the scale in order to obtain a stable two-factor structure (see Tables A.VI and A.VIII). Moreover, GPT 3.5 appears to differentiate between items that reference values and principles on the one hand and morality on the other hand in a way that humans and GPT 4 do not.

**Table A.V. GPT 4 Sample from December 2023– Initial Extraction and Final Solution for the Proximate Scale**

|  | Initial Extraction | | Final Extraction | |
|---|---|---|---|---|
|  | **CCM** | **MCM** | **CCM** | **MCM** |
| They disagree because they hold divergent opinions about the outcome of operating a day care center on company premises. | **0.892** | 0.018 | **0.885** | 0.03 |
| They disagree because they have different opinions about the effectiveness of opening a day care center for employees' children. | **0.809** | -0.022 | **0.811** | -0.029 |
| They disagree because they expect different consequences to follow from a company owned day care center. | **0.638** | 0.205 | **0.638** | 0.209 |
| They disagree because they have different views about how operating a day care center will impact the company. | **0.881** | -0.095 | **0.886** | -0.103 |
| They disagree about whether it is morally appropriate to open a day care center on company premises. | -0.051 | **0.913** | -0.052 | **0.91** |
| They disagree about whether it is morally acceptable for a company to offer day care for its employees' kids. | 0.001 | **0.967** | -0.005 | **0.95** |
| They disagree because they have incompatible views on whether operating a day care center is a righteous course of action. | 0.049 | **0.791** | 0.052 | **0.807** |
| They disagree because they support different principles on what a company should do for its employees. | -0.094 | 0.276 | ** | ** |

*Note*. Applied rotation method is oblimin. Factor loadings above 0.400 are indicated in bold. ** indicates item was eliminated prior to the presented solution.

**Table A.VI. GPT 3.5 Sample from December 2023 – Initial Extraction and Final Solution for the Proximate Scale**

|  | Initial Extraction | | Final Solution | |
|---|---|---|---|---|
|  | CCM | MCM | CCM | MCM |
| They disagree because they hold divergent opinions about the outcome of operating a day care center on company premises. | **0.665** | -0.105 | **0.683** | -0.112 |
| They disagree because they expect different consequences to follow from a company owned day care center. | **0.825** | 0.054 | **0.878** | 0.061 |
| They disagree because they have different views about how operating a day care center will impact the company. | **0.733** | -0.045 | **0.659** | -0.017 |
| They disagree because they have different opinions about the effectiveness of opening a day care center for employees' children. | 0.193 | -0.097 | ** | ** |
| They disagree about whether it is morally appropriate to open a day care center on company premises. | -0.022 | **0.985** | 0.021 | **1.073** |
| They disagree about whether it is morally acceptable for a company to offer day care for its employees' kids. | -0.07 | **0.87** | -0.048 | **0.802** |
| They disagree because they have incompatible views on whether operating a day care center is a righteous course of action. | 0.348 | **0.593** | ** | ** |
| They disagree because they support different principles on what a company should do for its employees. | 0.029 | 0.357 | ** | ** |

*Note.* Applied rotation method is oblimin. Factor loadings above 0.400 are indicated in bold. ** indicates item was eliminated prior to the presented solution.

**Table A.VII. GPT 4 Sample from December 2023 – Extraction Results for the Distal Scale**

|  | CCM | MCM |
|---|---|---|
| They disagree about the causes and consequences of their proposed actions. | **0.699** | -0.243 |
| They disagree about the consequences of their respective proposed actions. | **0.865** | -0.046 |
| They disagree because their views on the results of different courses of action are in contradiction. | **0.945** | 0.169 |
| They disagree on what will happen if their respective proposed actions are implemented. | **0.848** | -0.131 |
| They disagree because their views on which course of action is morally appropriate are in contradiction. | 0.029 | **0.823** |
| They disagree because they are incompatible in terms of their principles. | -0.165 | **0.694** |
| They disagree because they differ in their moral convictions. | -0.003 | **0.917** |
| They disagree because they have conflicting values. | -0.089 | 0.352 |

*Note.* Applied rotation method is oblimin. Factor loadings above 0.400 are indicated in bold.

**Table A.VIII. GPT 3.5 Sample from December 2023 – Initial Extraction and Final Solution for the Distal Scale**

|  | Initial Extraction | | | Final Solution | |
|---|---|---|---|---|---|
|  | F1 | F2 | F3 | CCM | MCM |
| They disagree about the causes and consequences of their proposed actions. | -0.11 | -0.208 | 0.334 | **0.387** | -0.122 |
| They disagree about the consequences of their respective proposed actions. | -0.237 | **0.429** | 0.242 | | |
| They disagree because their views on the results of different courses of action are in contradiction. | -0.004 | 0.229 | 0.319 | | |
| They disagree on what will happen if their respective proposed actions are implemented. | 0.042 | 0.011 | **0.939** | **0.787** | 0.021 |
| They disagree because their views on which course of action is morally appropriate are in contradiction. | 0.012 | **1.032** | -0.006 | | |
| They disagree because they are incompatible in terms of their principles. | **0.634** | 0.182 | -0.172 | -0.167 | **0.75** |
| They disagree because they differ in their moral convictions. | 0.283 | 0.183 | -0.256 | | |
| They disagree because they have conflicting values. | **1.024** | -0.04 | 0.068 | 0.103 | **0.887** |

*Note.* Applied rotation method is oblimin. Factor loadings above 0.400 are indicated in bold.

### A.2.3. Scale reliabilities by sample

All reliabilities calculated using four items for each scale.

**Table A.IX. Cronbach's alpha for each sample and scale**

|  | Pooled LLM | GPT 3.5 | GPT 4 | Human |
|---|---|---|---|---|
| Proximate Causal Misalignment Scale | 0.773 | 0.665 | 0.863 | 0.685 |
| Proximate Moral Misalignment Scale | 0.805 | 0.794 | 0.816 | 0.816 |
| Distal Causal Misalignment Scale | 0.735 | 0.468 | 0.928 | 0.820 |
| Distal Moral Misalignment Scale | 0.697 | 0.596 | 0.823 | 0.872 |

*Note.* Each scale is composed of 4 items.

### A.2.4. Diagnostic competence: *Comparison of means across conditions*

We calculate scale scores based on the hypothesized structure of the instrument to allow for comparisons with the human sample (Tables A.X and A.XI). Given the somewhat smaller sample size of GPT 3.5 responses as well as its performance in the above analyses, we focus our discussion on GPT 4, though report findings for GPT 3.5 as well.

GPT4 responses across the conditions exhibit human-like patterns of results with either version of the scale. Moral code disagreement attributions are higher in the moral misalignment condition (Distal: M = 3.24, SD = 0.63, Proximate: M = 2.36, SD = 0.48) compared to the causal misalignment condition (Distal: M = 2.07, SD = 0.39, Proximate: M = 1.63, SD = 0.21; $t_{distal}(48)=7.95$, $p_{distal} < 0.001$, $t_{proximate}(48)=7.00$, $p_{proximate} < 0.001$) and causal code disagreement attributions are higher in the causal code misalignment condition (Distal: M = 4.13, SD = 0.44, Proximate: M = 4.24, SD = 0.42) than in the moral misalignment condition (Distal: M = 3.54, SD = 0.57, Proximate: M = 3.82, SD = 0.29), $t_{distal}(48) = 4.11$, $p < 0.001$, $t_{proximate}(48)=4.12$, $p_{proximate} < 0.01$). In short, when comparing across conditions, mean attributions to misalignments in causal and moral codes reflect the experimental manipulation in the GPT 4 sample.

Table A.X. Mean comparisons across the two conditions for LLM and human samples on the proximate version of the causal and moral misalignment scales

|  | Causal Misalignment Condition | | | Moral Misalignment Condition | | |
| --- | --- | --- | --- | --- | --- | --- |
|  | Perceived moral code misalignment M (SD) | Perceived causal code misalignment M (SD) |  | Perceived moral code misalignment M (SD) | Perceived causal code misalignment M (SD) |  |
| LLM | 2.28 (0.90) | 3.91 (0.66) | p < .001 | 2.61 (0.77) | 3.62 (0.59) | p < .001 |
| GPT4 | 1.63 (0.21) | 4.24 (0.42) | p < .001 | 2.36 (0.48) | 3.82 (0.29) | p < .001 |
| GPT3.5 | 3.01 (0.81) | 3.55 (0.70) | p = .014 | 3.00 (0.98) | 3.31 (0.79) | p = .35 |
| Human | 2.10 (0.84) | 4.03 (0.64) | p < .001 | 3.53 (0.88) | 3.13 (0.88) | p = .02 |
|  |  |  |  |  |  |  |
| LLM vs Human | p = .34 | p = .43 |  | p < .001 | p = .002 |  |
| GPT4 vs Human | p = .009 | p = .144 |  | p < .001 | p < .001 |  |
| GPT3.5 vs Human | p < .001 | p = .008 |  | p = .04 | p = .46 |  |
| GPT3.5 vs GPT4 | p < .001 | p < .001 |  | p = .008 | p = .006 |  |

Table A.XI. Mean comparisons across the two conditions for LLM and human samples on the distal version of the causal and moral misalignment scales

|  | Causal Misalignment Condition | | | Moral Misalignment Condition | | |
| --- | --- | --- | --- | --- | --- | --- |
|  | Perceived moral code misalignment M (SD) | Perceived causal code misalignment M (SD) |  | Perceived moral code misalignment M (SD) | Perceived causal code misalignment M (SD) |  |
| LLM | 2.39 (.56) | 3.89 (.54) | p < .001 | 3.12 (.76) | 3.49 (.62) | p = .05 |
| GPT4 | 2.07 (.39) | 4.13 (.44) | p < .001 | 3.24 (.63) | 3.54 (.57) | p = .17 |
| GPT3.5 | 2.76 (.51) | 3.61 (.51) | p < .001 | 2.94 (.92) | 3.42 (.71) | p = .17 |
| Human | 2.15 (.77) | 4.08 (.73) | p < .001 | 3.54 (.89) | 3.26 (.92) | p = .09 |
|  |  |  |  |  |  |  |
| LLM vs Human | p = .10 | p = .16 |  | p = .02 | p = .16 |  |
| GPT4 vs Human | p = .62 | p = .77 |  | p = .13 | p = .16 |  |
| GPT3.5 vs Human | p = .002 | p = .01 |  | p = .02 | p = .52 |  |
| GPT3.5 vs GPT4 | p < .001 | p < .001 |  | p = .22 | p = .56 |  |

*A.2.5. Spillovers: Comparison of means within conditions*

Similar to human participants but to a higher degree, GPT 4 and GPT 3.5 see differences in causal codes to be at the root of moral misalignments as much as or more than differences in moral codes. Notably, while the spillover effect is not observed for humans when using the proximate scale, GPT 4 makes significantly higher attributions to causal code differences (M = 3.82, SD = 0.29) than to moral code differences (M = 2.36, SD = 0.48), $t(24) = 13.32$, $p < .001$) in the moral misalignment condition. Even though the condition manipulation results in the expected attributions, GPT 4 tends to see causal code misalignments as the primary underlying reason for either kind of disagreement.

*A.2.6. How discerning LLMs are: Comparisons of mean responses to the scale mean*

Given that we did not find scalar invariance in the MGCFA, we are not able to compare LLMs' and humans' mean scores. However, we compare groups' scores with the scale midpoint of 3 on a 5-point Likert scale to assess how discerning LLMs (and humans) are in separating causal and moral arguments.

In the moral misalignment condition, LLMs report moral misalignment below (for GPT 4 using a one-sided t-test: $t(24) = 6.68$, $p < .001$) or at (for GPT 3.5; $t(15) = 0$, $p = 1.00$) the scale midpoint for the proximate scale. When the distal scale is used, neither LLM sample's mean moral code misalignment attributions differ significantly from the scale midpoint (GPT 3.5: $t(15) = .272$, $p = .79$, GPT 4: $t(24) = 1.91$, $p = .07$). Taken together, these suggest that LLMs generally attribute disagreements to causal code misalignments even when moral misalignment is cued. One-sided t-tests indicate that causal code misalignments are significantly above the scale midpoint for GPT 4 on both scales (Proximate: $t(24) = 13.98$, $p < .001$, Distal: $t(24) = 4.76$, $p < .001$) and for GPT 3.5 when the distal scale is used ($t(15) = 2.39$, $p = .015$).

The results differ for the causal misalignment condition. For the proximate scale, mean attributions to causal code misalignments are above the scale mean for GPT 4 ($t(24) = 14.85$, $p < .001$) as well as GPT 3.5 ($t(21) = 3.67$, $p = .007$) on one-sided tests. Similarly, LLMs' attributions are significantly above the scale mean when we consider the distal scale (GPT 4: $t(24) = 12.85$, $p < .001$, GPT 3.5: $t(21) = 5.64$. $p < .001$). Moreover, in all cases, attributions to moral code disagreements are at or below the scale mean.

Overall, especially when using the proximate scale, GPT 4 has poor diagnostic performance in the moral misalignment condition due to under-estimating the extent of moral disagreement expressed in the conversation.

*A.2.7. Prevalence and patterns of mis-diagnoses: Cluster analyses*

In an average linkage hierarchical cluster analysis on the pooled data, using both the distal and proximate scales, we find two main clusters: 89 responses that have causal attributions greater than 3 ($M_{distal} = 3.37$, $M_{proximate}=3.22$) and higher moral attributions than causal attributions ($M_{distal}$ = 3.61, $M_{proximate}$ = 3.47) and 93 responses with high causal attributions ($M_{distal}$ = 3.94, $M_{proximate}$ = 4.05) and low moral attributions ($M_{distal}$ = 2.15, $M_{proximate}$ = 1.96). Tabulation of the former cluster across conditions show that 69 are in the moral misalignment condition but 20 are in the causal condition. Summarizing the scale means for these anomalous 20 observations, we find that responses to the distal scale are consistent with expectations ($M_{causal\_distal}$ = 3.79 and $M_{moral\_distal}$ = 3.1) but responses to the proximate scale violate expectations ($M_{causal\_proximate}$ = 3.37 and $M_{moral\_proximate}$ = 3.43). 6 of the 20 responses in this subset are from humans and 14 are from LLMs, all GPT 3.5. In other words, when using the proximate scale, GPT3.5 shows the same tendency that a small subset of humans do, to perceive moral code differences (in addition to causal code differences) in conversations that point to disagreements on causal codes.

In the second cluster making high causal attributions and low moral attributions, a sizable minority of 27 (out of 93) are in the moral misalignment condition. Both distal and proximate measures for this sub-sample show high attributions to causal code differences ($M_{causal\_distal}$ = 3.72, $M_{causal\_proximate}$ = 3.85) but low attributions to moral code differences ($M_{moral\_distal}$ = 2.43, $M_{moral\_proximate}$ = 2.26). Of these, 9 are human, 18 are LLM (6 are from GPT 3.5, 12 are from GPT4). This shows that LLMs have a higher tendency than humans and GPT 4 has a higher tendency than GPT 3.5 to make causal attributions and not recognize differences in moral codes in the moral misalignment condition.

A separate analysis using only means of the distal scales for causal and moral misalignment creates three main clusters but points to a similar set of findings. One cluster of 15 responses, all in the moral misalignment condition, registers high moral attributions ($M_{moral\_distal}$ = 3.97) and low causal attributions ($M_{causal\_distal}$ = 2.18), as we expected. 10 of these are from humans and 5 are from LLMs (3 GPT3.5, 2 GPT 4).

A second cluster of 72 observations register high levels of both causal ($M_{causal\_distal}$ = 3.67) and moral attributions ($M_{moral\_distal}$ = 3.61). Of these, 18 are in the causal misalignment condition and 54 are in the moral misalignment condition. The former clearly perceive causal misalignment ($M_{causal\_distal}$ = 3.81) but also perceive a non-trivial level of moral code differences ($M_{moral\_distal}$ = 3.19, SD=.44, t(17) = 1.87, p=0.04 on one-tailed t-test of difference from 3). 8 of these responses are from humans and 10 from LLMs, all GPT 3.5. These findings show, that GPT 3.5's tendency (along with a small subset of humans) to perceive moral code differences (in addition to causal code differences) in conversations that point to disagreements on causal codes is evident on the distal scale as well.

That said, spillovers in attributions are more likely in the moral misalignment condition. 54 observations in this condition register high levels of moral code disagreement ($M_{moral\_distal}$ = 3.75) but also high levels of causal code disagreement ($M_{causal\_distal}$ = 3.62). 34 are human, 20 are LLM (6 GPT 3.5, 14 GPT 4) responses.

The third cluster of 91 responses shows high causal attributions and low moral attributions. 24 of these are in the moral misalignment condition, pointing to an anomaly. 8 of these were given by human participants and 16 by LLMs (7 GPT 3.5, 9 GPT 4). Considering that there are 56 humans and 41 LLM responses in the moral misalignment condition, we conclude that LLMs are more than twice as likely as humans to make causal attributions and not recognize moral code differences in the moral misalignment condition.

Overall, cluster analyses show that LLMs have a greater tendency than humans to make unexpected attributions, especially in the moral misalignment condition. Even on the distal scale, where we find human-like performance on average for GPT 4 in the previous analyses, only 2 out of 25 GPT 4 responses assigned to the moral misalignment condition make appropriately high moral attributions and low causal attributions. The equivalent proportion is 10 out of 56 for humans. 14 of the 25 (compared to 34 of the 56 human responses) register high causal disagreement along with high moral disagreement. 9 of the 25 (compared to 8 of 56 human responses) make high attributions to causal code differences only, reporting low levels of moral code disagreement.

We find that responses that run counter to expectations are not unique to LLMs – small subsets of human participants show the same tendencies. However, LLMs are more likely than humans to mis-diagnose sources of conflict reflected in conversations. First, LLMs are more than twice as likely as humans to make causal attributions and not recognize moral code differences in the moral misalignment condition. GPT 4 has a higher tendency than GPT 3.5 to make this misattribution. Even on the distal scale, where we find human-like performance on average for GPT 4 in the previous analyses, only 2 out of 25 GPT 4 responses assigned to the moral misalignment condition make appropriately high moral attributions and low causal attributions. The equivalent proportion is 10 out of 56 for humans. 14 of the 25 (compared to 34 of the 56 human responses) register high causal disagreement along with high moral disagreement. 9 of the 25 (compared to 8 of 56 human responses) make high attributions to causal code differences only, reporting low levels of moral code disagreement.

Second, when using either the proximate or the distal scale, GPT 3.5 shows the same tendency that a small subset of humans do, to perceive moral code differences (in addition to causal code differences) in conversations that point to disagreements on causal codes. Thus, while GPT 4 is likely to err on the side of making causal attributions, GPT 3.5 makes both kinds of misattributions – over-attributing to moral code differences when the disagreement is rooted in causal codes in addition to over-attributing to causal code differences when the disagreement is rooted in moral codes.

*A.2.8. Correlations of misalignment and intragroup conflict*

We present correlations between dimensions of intragroup conflict and both versions of the code misalignment scale in Tables A.XII and A.XIII. The distal and proximate versions of each code misalignment scale exhibit large and significant correlations as expected. However, correlation coefficients for the moral code misalignment scale are markedly lower for the LLM samples (GPT 3.5: 0.51, $p < 0.05$, GPT 4: 0.68, $p < 0.05$) compared to the human sample (0.82, $p < 0.05$). Moreover, the pattern of correlations differs in several cases across samples. First, Kocak et al (2023) found relationship conflict to correlate significantly (and negatively) with the likelihood of resolution in the human sample ($\rho = -0.20$, $p < 0.05$). However, none of the intragroup conflict measures exhibit significant correlations with likelihood of conflict resolution for the LLM subsamples. Second, we find a significant and positive correlation between likelihood of conflict resolution and proximate causal code misalignment attributions (ρ = .46, p < .05) for the GPT 3.5 sample but not in the human or GPT 4 samples. Finally, whereas human responses tend to show positive correlations between code misalignment and conflict, neither GPT 4 nor GPT 3.5 responses show these, suggesting that unlike humans LLMs do not associate disagreements with conflicts.

**Table A.XII. GPT3.5 Sample from December 2023**

| GPT 3.5 SAMPLE | I | II | III | IV | V | VI | VII |
|---|---|---|---|---|---|---|---|
| I. Distal moral CM | 1 | | | | | | |
| II. Distal causal CM | -0.11 | 1 | | | | | |
| III. Proximate moral CM | 0.51* | -0.05 | 1 | | | | |
| IV. Proximate causal CM | -0.22 | 0.57* | 0.09 | 1 | | | |
| V. Task conflict | 0.04 | -0.06 | 0.08 | 0.10 | 1 | | |
| VI. Relationship conflict | 0.05 | 0.07 | -0.07 | 0.04 | 0.29 | 1 | |
| VII. Process conflict | -0.17 | 0.05 | -0.03 | 0.15 | 0.74* | 0.35* | 1 |
| VIII. Likelihood of conflict resolution | -0.27 | 0.09 | -0.03 | 0.46* | 0.06 | -0.03 | 0.17 |

* p < .05

**Table A.XIII. GPT 4 Sample from December 2023**

| GPT 4 SAMPLE | I | II | III | IV | V | VI | VII |
|---|---|---|---|---|---|---|---|
| I. Distal moral CM | 1 | | | | | | |
| II. Distal causal CM | -0.64* | 1 | | | | | |
| III. Proximate moral CM | 0.68* | -0.49* | 1 | | | | |
| IV. Proximate causal CM | -0.66* | 0.68* | -0.31* | 1 | | | |
| V. Task conflict | -0.05 | 0.11 | -0.05 | -0.03 | 1 | | |
| VI. Relationship conflict | -0.02 | 0.12 | -0.14 | 0.08 | 0.13 | 1 | |
| VII. Process conflict | 0.10 | -0.03 | 0.15 | -0.10 | 0.51* | 0.22 | 1 |
| VIII. Likelihood of conflict resolution | -0.18 | 0.16 | -0.17 | 0.14 | 0.01 | 0.17 | 0.08 |

* p < .05